\documentclass[runningheads]{llncs}
\usepackage[T1]{fontenc}
\usepackage{graphicx}
\usepackage{booktabs}
\usepackage[misc]{ifsym}
\newcommand{\corr}{(\Letter)}

\usepackage{amsmath}
\usepackage{amssymb}
\usepackage{comment}
\usepackage{euflag}
\usepackage{tcolorbox}
\usepackage{algorithm}
\usepackage{hyperref}
\usepackage{multirow}
\usepackage{xcolor} % Required for 'green' and color management
\usepackage{soul}   % Required for \sethlcolor and \hl
\usepackage{orcidlink}
\usepackage{algpseudocode}

\usepackage{tikz}
\usepackage{float}
\newcommand{\chill}{\textsf{CHiL(L)Grader}}
\definecolor{dotblue}{HTML}{0057D2}
\newcommand{\bluedot}[1]{\resizebox{!}{10pt}{
\begin{tikzpicture}[baseline=-4pt]
    \draw[dotblue,fill=dotblue] (0,0) circle (5pt);
    \draw (0,0) node {\normalsize\textsf{\color{white} #1}};
\end{tikzpicture}
}}

\begin{document}

\title{\chill{}: \underline{C}alibrated \underline{H}uman-\underline{i}n-the-\underline{L}oop Short-Answer Grading}

\author{Pranav Raikote\inst{1}\orcidlink{0009-0008-7464-5828}\corr\and
Korbinian Randl\inst{1}\orcidlink{0000-0002-7938-2747} \and
Ioanna Miliou\inst{1}\orcidlink{0000-0002-1357-1967} \and \\ Athanasios Lakes\inst{1}\orcidlink{0009-0005-4803-4722} \and Panagiotis Papapetrou\inst{1}\orcidlink{0000-0002-4632-4815}}

\authorrunning{P. Raikote et al.}
\institute{
    Department of Computer and Systems Sciences, Stockholm University, \\
    164 25 Kista, Sweden \\
    \email{\{pranav.raikote,korbinian.randl,ioanna.miliou, \\athanasios.lakes,panagiotis\}@dsv.su.se}}

\maketitle              

\begin{abstract}
Scaling educational assessment with large language models requires not just accuracy, but the ability to recognize when predictions are trustworthy. Instruction‑tuned models tend to be overconfident, and their reliability deteriorates as curricula evolve, making fully autonomous deployment unsafe in high‑stakes settings. We introduce \chill{}, the first automated grading framework that incorporates calibrated confidence estimation into a human‑in‑the‑loop workflow. Using post‑hoc temperature scaling, confidence‑based selective prediction, and continual learning, \chill{} automates only high‑confidence predictions while routing uncertain cases to human graders, and adapts to evolving rubrics and unseen questions. Across three short answer grading datasets, \chill{} automatically scores $35-65\%$ of responses at expert‑level quality (QWK $\ge 0.80$). A QWK gap of $+0.347$ between accepted and rejected predictions confirms the effectiveness of the confidence‑based routing. Each correction cycle strengthens the model's grading capability as it learns from teacher feedback. These results show that uncertainty quantification is key for reliable AI‑assisted grading.

\end{abstract}

\section{Introduction}
\label{sec:introduction}
\begin{figure}[ht]
    \centering
    \includegraphics[width=0.9\linewidth]{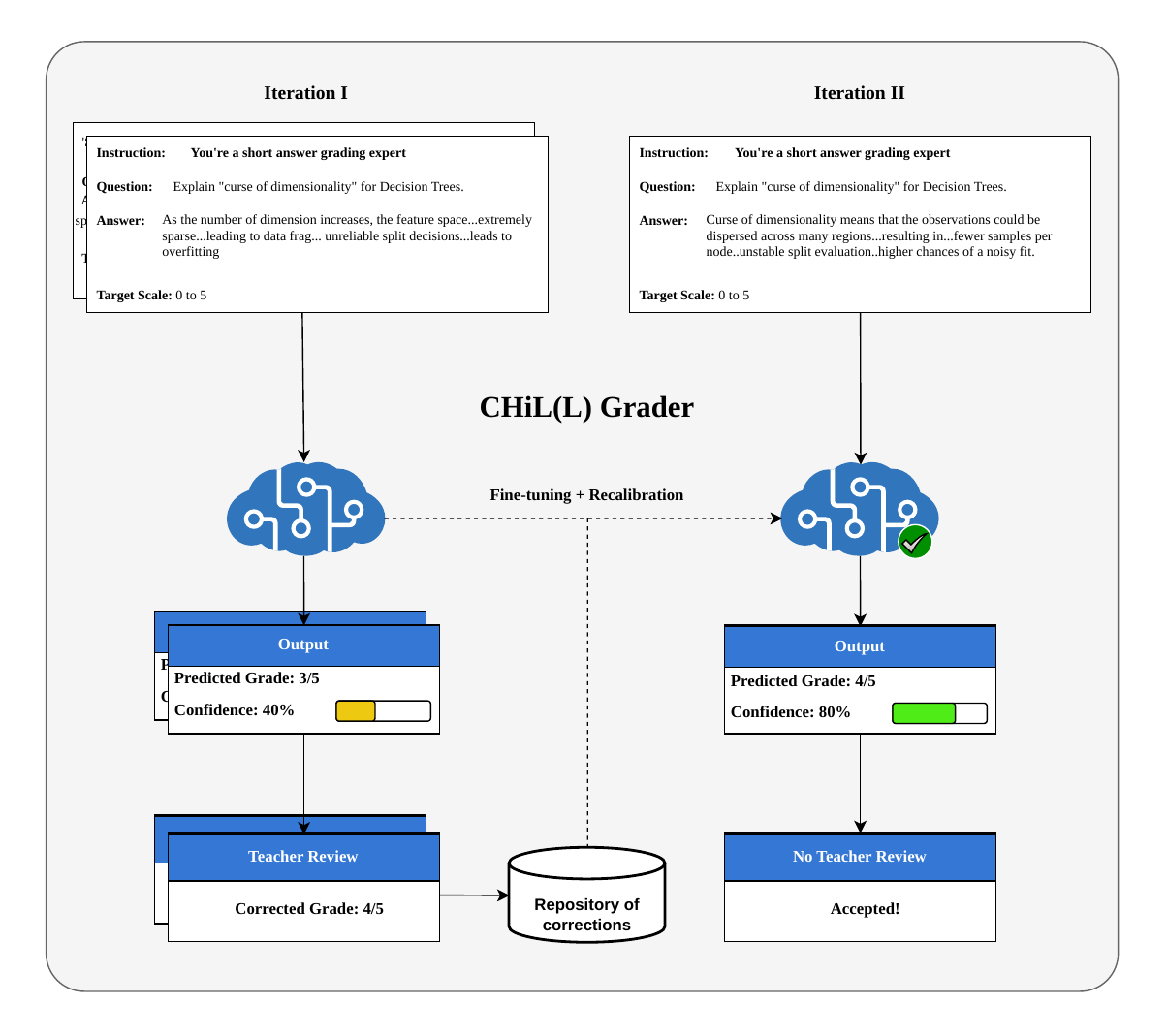}
    \caption{The \chill{} loop over two iterations for similar responses to the same question. In Iteration I, the model predicts 3/5 with low confidence (40\%), triggering teacher review; the corrected 4/5 grade is used for fine‑tuning. In Iteration II, the updated model predicts 4/5 with high confidence (80\%), enabling automatic acceptance.}
    \label{fig:overview}
\end{figure}
Education systems worldwide are rapidly transforming due to the expansion of higher education and the rise of online and blended learning environments~\cite{marginson2016worldwide}. While this growth improves access to education, it also increases pressure on instructional resources, as educators must deliver high-quality teaching, timely feedback, and fair assessment to larger and more diverse student cohorts. In response, Machine Learning~(ML) has emerged as a promising means to support teaching and learning, through the adoption of ML-based tools, such as intelligent tutoring systems, adaptive feedback mechanisms, and personalized learning environments~\cite{feng2023aesqwk}. These developments reflect a broader shift toward data-driven, AI-supported education, further accelerated by advances in Large Language Models~(LLMs). With strong capabilities in reasoning, summarization, and evaluation, LLMs have increased interest in applying ML to core educational applications, including automated assessment and feedback generation. 

Among these applications, assessment stands out as particularly impactful, since it shapes learning trajectories, determines academic progression, and directly affects student outcomes. Nonetheless, large-scale assessment faces a persistent trade-off between quality and feasibility. Human grading ensures accuracy and personalized feedback, but becomes costly and time-consuming~\cite{burrows2015eras}. LLMs offer a compelling alternative for Automated Short-Answer Grading~(ASAG) tasks, demonstrating strong performance across various educational domains~\cite{aggarwal2025understand,Impey_2025,jimmy2021asag}. However, despite their competitive performance, their deployment in high-stakes educational settings remains constrained by two fundamental challenges:

\vspace{4pt}
\noindent\textbf{(i)}~\textbf{LLMs exhibit systematic overconfidence} in their predictions~\cite{geng2024survey,guo2017calibration,xiong2024llms}. They routinely assign high confidence scores even to incorrect predictions, providing no reliable mechanism for teachers to determine when model outputs can be trusted and when human intervention is necessary. This miscalibration problem is particularly acute in educational contexts where the consequences of grading errors affect academic progression.

\vspace{4pt}
\noindent\textbf{(ii)}~\textbf{Model performance degrades substantially under distribution shi-ft}~\cite{ovadia2019trust}, as systems encounter different question types, grading rubrics, and response patterns from those seen during initial training. 
This brittleness to distributional changes is especially problematic in educational settings where the curriculum evolves and instructors modify questions and grading rubrics~\cite{burrows2015eras}.

These limitations prevent fully autonomous ASAG systems from being deployed in educational contexts. Effective use requires methods that reduce instructor effort, adapt over time, and produce calibrated, trustworthy confidence scores. Human-in-the-Loop (HiL) frameworks improve reliability by combining automated grading with human oversight. In this paper, we address the above limitations by proposing a calibrated HiL framework for ASAG that integrates three complementary mechanisms: \textbf{post-hoc calibration} to obtain reliable confidence estimates~\cite{guo2017calibration}, \textbf{selective prediction} to defer uncertain cases to human review, and a \textbf{continual learning loop} that incorporates human feedback to adapt to new grading conditions while mitigating catastrophic forgetting~\cite{delange2022continual,wang2023orthogonal}. 
As illustrated in Figure~\ref{fig:overview}, the LLM predicts both a grade and an associated confidence score for each exam question response; high-confidence predictions are accepted automatically, while low-confidence cases are reviewed and corrected by human evaluators. These corrected samples are accumulated to iteratively retrain and recalibrate the model. 
In summary, our contributions are as follows:
\begin{enumerate} 
    \item \textbf{Novelty.} We introduce \chill, the first HiL framework for ASAG that integrates (i) confidence calibration, (ii) confidence-based selective prediction, and (iii) principled human deferral into a unified system. It achieves expert-level grading performance (QWK $\geq 0.80$) on at least $30\%$ (and up to $65\%$) of student responses, while systematically deferring uncertain cases to human review.
    \item \textbf{Reliability.} \chill{} addresses the overconfidence challenge by introducing temperature scaling based on the Expected Calibration Error~(ECE) \cite{guo2017calibration,posocco2021calibrate} and empirically demonstrating consistent reductions across three data-sets and yielding calibrated confidence estimates suitable for selective prediction. 
    \item \textbf{Adaptation.} We demonstrate that HiL-based continual learning enables generalization under distribution shift. By leveraging human corrections as supervision, our approach maintains consistent improvements in performance across questions and rubrics that differ from the initial training conditions.
    \item \textbf{Effectiveness. } Our experiments on three datasets spanning different domains, grading scales, and difficulty levels show that \chill{}  consistently improves the grading quality, while reducing the number of required manual corrections. 
    \item \textbf{Reproducibility.} Our code, model configurations, and evaluation scripts are publicly available on GitHub \footnote{https://anonymous.4open.science/r/chil-grading-96A3/README.md}.
\end{enumerate}

\section{Related Work}
Early ASAG systems~\cite{Burstein1999,Leacock2003CraterAS}  relied on domain-specific semantic representations and rule-based concept matching to identify key ideas in student responses. Later work introduced statistical and semantic-similarity approaches, including vector-space and regression models for ASAG~\cite{dzikovska-etal-2013-semeval,mohler-mihalcea-2009-text}. More recently, LLMs have demonstrated strong performance for ASAG, for example by showing that GPT-4, when prompted with structured rubrics, can produce scores comparable to teachers and outperform peer grading~\cite{Impey_2025}. Recent advances show that LLMs can achieve teacher‑level performance when guided by structured rubrics~\cite{Impey_2025}, and fine‑tuned models such as LLaMA‑2 and Mistral can generate richer grading feedback~\cite{aggarwal2025understand}. For longer responses, GPT‑3.5‑based feedback generation and targeted fine‑tuning have further improved beyond simple score prediction~\cite{zeinalipour2025}.

Several works enhance factual grounding in ASAG using Retrieval Augmented Generation (RAG).  For instance, Duong et al.~\cite{duong2024asag} retrieve embedded lecture notes to guide GPT-3.5/4 grading, improving correlation with human evaluators, while Chu et al.~\cite{chu2025enhancingllmbasedshortanswer} build a multi index knowledge base combining course materials and graded examples, yielding consistent zero shot gains. These approaches enhance accuracy and contextual grounding but largely assume fully automated grading.  Human preference integration has largely focused on offline alignment. RLHF has been applied using Stack Overflow votes to fine tune GPT Neo~\cite{gorbatovski2024}, and DPO has been used to optimize feedback quality based on teacher preferences in classroom settings~\cite{2025.EDM.short-papers.166}. While these methods align models with human judgments, they rely on static fine tuning and do not incorporate uncertainty estimation or mechanisms for deferring uncertain cases.

More recently interactive workflows and confidence estimation have gained interest. Systems, such as GradeHITL~\cite{chu2025enhancingllmbasedshortanswer} and Avalon~\cite{avalon}, involve instructor-in-the-loop mechanisms to improve rubric alignment. In parallel, neural models are often overconfident in their predictions and can be improved by post-hoc methods like temperature scaling ~\cite{guo2017calibration}, with extensions to LLMs, such as Adaptive Temperature Scaling~\cite{xie2024calibrating}.  While reliable confidence is essential for selective prediction and human deferral, these calibration methods have not yet been integrated into ASAG pipelines that route uncertain responses to human graders.

Although prior work has improved grading accuracy, rubric alignment, and calibration in isolation, it has not yet combined calibrated confidence into deploy-ment-time HiL deferral for ASAG; this gap motivates our approach.

\section{Problem Formulation and Framework}

\subsection{Problem Formulation}
In this paper, we focus on short-answer grading, assessing responses for correctness in alignment with a given grading rubric. Let ${\bf q}$ be an exam question, ${\bf a}$ the corresponding student answer, and $\mathcal{G}$ the grading rubric, defined as follows:
\begin{definition}[Rubric]\label{def:rubric}
Let $G \in \mathbb{Z}_{\ge 0}$ denote the maximum attainable grade under a given grading scheme.
A \emph{rubric}~$\mathcal{G}$ is defined as the set of all possible grades permitted under that scheme:
\begin{equation}
\mathcal{G} = \{0, 1, \ldots, G\} .
\end{equation}
In our formulation we assume an integer-based grading scale, but our solution can extend to any grading scale by applying discretization.
\end{definition}
Our objective is to train a classifier~$f(\cdot)$ that predicts a \emph{grade}~$\hat{g}\in \mathcal{G}$ for a given %\emph{ground truth grade}~$g$, 
\emph{exam question}~${\bf q}$,  \emph{student answer}~${\bf a}$, and rubric $\mathcal{G}$:
\begin{equation}\label{eq:problem1}
    \hat{g} = f({\bf q}, {\bf a}, \mathcal{G}).
\end{equation}
Both texts are represented as sequences of tokens from a vocabulary~$\mathcal{V}$, i.e., ${\bf q} = (q_1, \ldots,$ $q_{|{\bf q}|})$, ${\bf a} = (a_1, \ldots, a_{|{\bf a}|})$, 
where $q_i, a_j \in \mathcal{V}$ denote the $i$-th question token and the $j$-th answer token, respectively, and $|{\bf q}|$ and $|{\bf a}|$ are the numbers of tokens in the question and answer texts.

Next, we formalize two challenges of ASAG (as also indicated in Sec. \ref{sec:introduction}). Firstly, $f(\cdot)$ tends to be overconfident in its predictions.

\begin{definition}[\textbf{Overconfidence}]
Let $\mathbb{P}_f({\bf q}, {\bf a}, \mathcal{G})$ be the predictive distribution over grades $g^* \in \mathcal{G}$ induced by $f(\cdot)$. The predicted grade $\hat g$ and its confidence $\hat c$ are defined as $\hat g = \arg\max \mathbb{P}_f({\bf q}, {\bf a}, \mathcal{G})$ and $\hat c = \max \mathbb{P}({\bf q}, {\bf a}, \mathcal{G})$, respectively. 
$f(\cdot)$ is \emph{overconfident} if 
\begin{equation}
\mathrm{P}(\hat g = g \mid \hat c = c) < c.
\end{equation}
\end{definition}
Moreover, the performance of $f(\cdot)$ drops under distribution shift.

\begin{definition}[\textbf{Distribution Shift}]
Let $\mathbb{P}_S({\bf q},{\bf a},g)$ and $\mathbb{P}_T({\bf q},{\bf a},g)$ denote the training and test distributions. 
A \emph{distribution shift} occurs when $\mathbb{P}_S \neq \mathbb{P}_T$, for instance due to changes in question types, grading rubric $\mathcal{G}$, or student response patterns. The resulting degradation in model performance is measured by the increase in expected grading loss $l_f(\cdot)$:
\begin{equation}
\Delta_{\text{shift}}(f)
=
\mathbb{E}_{\mathbb{P}_T}\!\left[\ell_f(\hat g,g)\right]
-
\mathbb{E}_{\mathbb{P}_S}\!\left[\ell_f(\hat g,g)\right]
> 0.
\end{equation}
\end{definition}
Given the set $\mathcal{F}$ of acceptable target grading functions, the two problems studied in this paper are as follows:
\begin{problem}[\textbf{Min-max confidence gap}]
\label{prob:minmax}
Learn a function~$f(\cdot)$, such that the maximum confidence gap is minimized:
\begin{equation}
\min_{f\in\mathcal{F}}\ \max_{c\in[0,1]}\Big|\mathrm{P}(\hat g=g \mid \hat c= c)-c\Big|.
\end{equation}
\end{problem}
 
\begin{problem}[\textbf{Minimum grading error under distribution shift}]
\label{prob:shift}
Given a source distribution $\mathbb{P}_S$ and a shifted target distribution $\mathbb{P}_T\neq\mathbb{P}_S$, learn $f(\cdot)$ that minimizes the target grading error:
\begin{equation}
\min_{f\in\mathcal{F}}\ \mathbb{E}_{(X,g)\sim\mathbb{P}_T}\big[\ell(\hat g_f(X),g)\big].
\end{equation}
\end{problem}

\subsection{Our \chill{} Framework}

\begin{figure}[t]
\includegraphics[width=\textwidth]{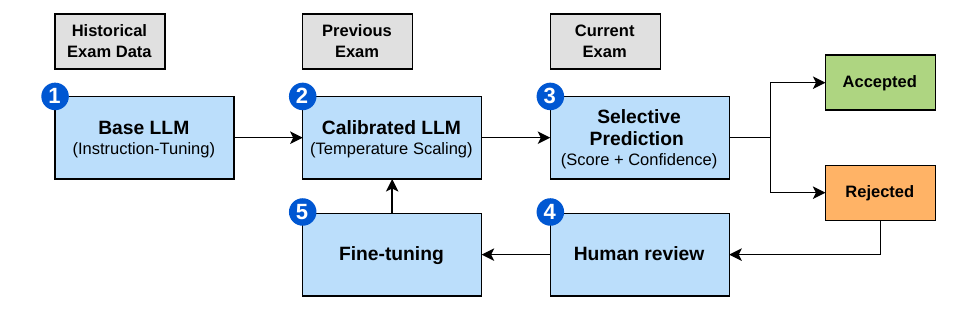}
\caption{\chill{} architecture. Historical exams are used to train the instruction‑tuned model. Prior‑year exams calibrate its confidence. During the current exam, low‑confidence cases are sent to human review, whose corrections, combined with replay samples, guide conservative model updates and recalibration.} 
\label{fig1:hil_pipeline}
\end{figure}

The overall architecture of the \chill{} framework is illustrated in Figure~\ref{fig1:hil_pipeline}.
Specifically,\bluedot{1}we instruction-tune a text-to-text LLM for ASAG, enabling it to faithfully follow grading rubrics and maintain consistent in-domain performance~\cite{zhang2025it}.
Then\bluedot{2}we employ a \textbf{post-hoc calibration technique}, i.e., \textbf{temperature scaling}, to transform unreliable model confidence scores into well-calibrated probability estimates that accurately reflect true prediction uncertainty~\cite{guo2017calibration}. Next,\bluedot{3}we implement \textbf{selective prediction with explicit coverage}, allowing the model to handle only high-confidence cases, while routing uncertain instances to human review. This selective routing ensures that human expertise is allocated where model predictions are least reliable, rather than being uniformly distributed across all grading decisions. Finally,\bluedot{4}\textbf{+}\bluedot{5}we introduce a \textbf{continual learning loop}, in which human corrections serve as high-quality supervision. These corrections, combined with a replay buffer~\cite{wang2023orthogonal} to mitigate catastrophic forgetting~\cite{delange2022continual}, enable the model to adapt to new question types and grading criteria while preserving previous knowledge.

This creates a loop where human expertise continuously refines model behavior, progressively expanding the scope of safe automation while maintaining grading quality.
Although grading is inherently a classification problem over the discrete rubric set 
$\mathcal{G}$, our approach employs a generative LLM. Thus, we formulate the task described in Eq. \ref{eq:problem1} as a sequence-to-sequence generation problem, where the model produces a structured output string encoding the predicted grade
$\hat{\bf g} = f({\bf q}, {\bf a}, \mathcal{G})$, with the output vector~$\hat{\bf g}$ being a tokenized structured string of the form \texttt{\{"grade": $\hat{g}$, "max\_grade": $G$\}}.
The full procedure is summarized in Algorithm~\ref{algo:hil_grading}. In the following, we explain each step in detail:

\begin{algorithm}[t]
\caption{Calibrated Human-in-the-Loop ASAG}
\label{algo:hil_grading}
\begin{algorithmic}[1]

\Statex \textbf{Input:} training set $\mathcal{D}_\text{train}$, calibration set $\mathcal{D}_\text{cal}$, test splits $\{D_{21}, \ldots, D_{2N}\}$, confidence threshold $\tau \in [0,1]$, replay buffer size $k$.
\Statex \textbf{Output:} grading model $f_\theta$ and calibrated temperature $T^*$.
\Statex
\Statex \textbf{Stage 1: Instruction-tuning and Calibration}
\State Train $f_\theta$ on $\mathcal{D}_\text{train}$

\For{$({\bf q}, {\bf a}, g, \mathcal{G}) \in \mathcal{D}_\text{cal}$}
    \State $z_{g^*} \gets \log \mathrm{P}(g^* \mid {\bf q}, {\bf a}, \mathcal{G}), \quad \forall g^* \in \mathcal{G}$ \hfill\bluedot{1}
\EndFor

\State $T^* \gets \arg\min_{T}\;\text{ECE}\!\left(
    \mathrm{softmax}({\bf z}/T),\, g\right), \quad g \in 
    \mathcal{D}_\text{cal}$ \hfill\bluedot{2}

\Statex
\Statex \textbf{Stage 2: Human-in-the-Loop Continual Learning}
\State $\mathcal{H} \gets \emptyset$
\For{$j = 1, \ldots, N$}
    \For{each $({\bf q}, {\bf a}, \mathcal{G}) \in D_{2j}$}
        \State ${\bf p} \gets \mathrm{softmax}({\bf z}({\bf q}, 
        {\bf a}, \mathcal{G}) / T^*)$
        \State $\hat{g} \gets \arg\max\; {\bf p}$;\quad 
               $\hat{c} \gets \max\; {\bf p}$ \hfill\bluedot{3}
        \If{$\hat{c} \geq \tau$}
            \State \textbf{accept} $\hat{g}$ as final grade
        \Else
            \State \textbf{reject}; obtain human correction $\bar{g}$ \hfill\bluedot{4}
            \State $\mathcal{H} \gets \mathcal{H} \cup \{({\bf q}, {\bf a}, \bar{g}, \mathcal{G})\}$
        \EndIf
    \EndFor
    \State Construct $\mathcal{B}$: retrieve $k$ similar questions from $\mathcal{D}_\text{train}$ per question in $\mathcal{H}$
    \State Fine-tune LoRA adapters of $f_\theta$ on $\mathcal{H} \cup \mathcal{B}$ \hfill\bluedot{5}
    \State Recalibrate $T^*$ on accepted samples from $\mathcal{D}_{2j}$
\EndFor

\end{algorithmic}
\end{algorithm}

\smallskip
\noindent
\textbf{\bluedot{1}Instruction-Tuning}
Instruction-tuning adapts a pretrained LLM to follow task-specific instructions on structured prompt-response pairs enabling the model to follow the grading rubrics and produce consistent grade predictions~\cite{zhang2025it}. Our training data consists of question–answer pairs with human‑provided grades (see Section~\ref{sec:data}), formatted using standardized prompts that explicitly reference $\mathcal{G}$. Some training examples are shown in the Appendix.

\smallskip
\noindent
\textbf{\bluedot{2}Model Calibration}
Instruction-tuned models are heavily overconfident: they assign confidence scores near $0.99$ even to incorrect predictions, rendering raw confidence unreliable for deferring decisions. We address this through \textbf{post-hoc temperature scaling}.
At inference, we deterministically score all possible grades to avoid sampling variability. The model’s log likelihood to predict a grade $g\in\mathcal{G}$ is obtained by conditioning on the corresponding structured response and summing the token-level log probabilities:
\begin{equation}
{\bf z} = \left[ \left. \sum_{i=1}^{|{\bf g}|} \log \mathrm{P}\big(t_{i}\mid {\bf q}, {\bf a}, \mathcal{G}, {\bf g}_{<i}\big) ~\right|~ \begin{array}{rl}
\small\forall{\bf g} = \text{\tt \{} & \small\text{\tt "grade": ${g^*}$,}\\ 
\small & \small\text{\tt "max\_grade": $G$~\}}
\end{array}; g^* \in \mathcal{G} \right ].
\label{eq:candidate_scoring}
\end{equation}
Here $t_i \in \mathcal{V}$ is the $i$-th token in ${\bf g}$, ${\bf g}_{<i}$ is the vector of tokens preceding $t_i$, and $|{\bf g}|$ is the  length of ${\bf g}$.

Temperature scaling rescales the elements of this logit vector ${\bf z} = [z_0, z_1, ..., z_{G}]$ by a single learned scalar $T > 0$ to produce calibrated class probabilities: 
\begin{equation} 
    \mathrm{P}(g^*;~T) = \frac{e^{(z_{g^*} / T)}}{\sum_{h \in \mathcal{G}} e^{(z_h / T)}}. 
\label{eq:temperature_scaling} 
\end{equation} 
The predicted grade~$\hat{g}$ and its confidence~$\hat{c}$ are then:
\begin{align} 
\hat{g}=\arg\max_{g^*\in\mathcal{G}}\mathrm{P}(g^*;T),
\qquad
\hat{c}=\max_{g^*\in\mathcal{G}}\mathrm{P}(g^*;T).
\end{align}

Miscalibration is quantified by ECE~\cite{guo2017calibration}, which partitions predictions into $B$ equal-width confidence bins and measures the weighted gap between predicted confidence and empirical accuracy.
\begin{definition}[Expected Calibration Error]
    Let $n_b$ denote the number of observations in bin $b$, while $\mathrm{P}_b(\cdot)$ and $\mathbb{E}_b[\cdot]$ are the statistical probability and the expected value based on observations in bin~$b$. Then ECE is computed as:
    \begin{equation}
        \text{ECE} = \sum_{b=1}^{B} \frac{n_b}{n} \bigl|\,\mathrm{P}_b(\hat{g} = g) - \mathbb{E}_b[\hat{c}]\bigr|.
        \label{eq:ece}
    \end{equation}
\end{definition}
ECE $= 0$ denotes perfect calibration while ECE $< 0.1$ is considered acceptable in practice~\cite{nixon2020measuring}.
The optimal temperature is found by sweeping $T \in [0.1, 2.0]$ on a held-out calibration split to minimize ECE
\begin{equation} 
    T^* = \arg\min_{T}\;\text{ECE}\,(\mathrm{softmax}({\bf z}/T), g),
    \label{eq:temp}
\end{equation} 
where $g \in \mathcal{D}_\text{cal}$ denotes the ground-truth labels from the held-out calibration set. Temperature scaling preserves predictive accuracy while producing confidence scores that accurately reflect the model's true reliability.

\smallskip
\noindent
\textbf{\bluedot{3}Selective Prediction}
After calibration, each prediction is routed through a confidence gate with threshold $\tau$. Given a student response, the model produces a calibrated confidence $\hat{c} \in [0,1]$. If $\hat{c} \ge \tau$, the prediction is \emph{accepted} as the final grade, otherwise it is \emph{rejected} and sent to a teacher for manual grading. The threshold $\tau$ is chosen by performing a post-hoc sweep over a held-out, fully-graded calibration set, evaluating both accuracy and coverage of the accepted subset. We select the largest $\tau$ that satisfies a pre-specified reliability target on the accepted set. This yields an operating point such that only predictions that are both high-confidence and reliable are accepted, while the others are systematically redirected to human graders.

\smallskip
\noindent
\textbf{\bluedot{4}Human Review} Low-confidence predictions are reviewed by human graders, whose corrections form the HiL set $\mathcal{H}$. Each correction provides a human-verified grade for each rejected sample, offering targeted supervision that addresses the model's specific weaknesses under the current exam's grading conditions.

\smallskip
\noindent
\textbf{\bluedot{5}Replay Augmented Fine-tuning}
To preserve prior performance and stabilize adaptation across heterogeneous scoring scales, we augment $\mathcal{H}$ with a replay buffer $\mathcal{R}$ comprising stratified samples from historical training data. The replay buffer is \emph{scale-aware}: for each maximum attainable grade $G \in \mathbb{Z}_{\ge 0}$ represented in $\mathcal{H}$, we retrieve a balanced set of historical examples, such that the grade-scale distribution over $\mathcal{H} \cup \mathcal{R}$ mirrors that of the corrections. This prevents overfitting to the dominant scales of the current exam and anchors learning to the broader historical distribution, preserving consistent performance across question types.

We fine-tune only the adapters from a Low-Rank Adaptation~(LoRA) process~\cite{hu2022lora} on $\mathcal{H} \cup \mathcal{R}$, keeping the base model weights and prompts fixed. Adapter-only updates allow the model to adjust to unseen question-answer patterns and evolving grading styles without degrading previously acquired capabilities. By interleaving corrections with diverse historical samples spanning multiple question types, rubric structures, and difficulty levels, the replay buffer serves as a regularizer against catastrophic forgetting~\cite{kotha2024understanding,luo2025forget}. After each fine-tuning step, temperature $T^*$ is recalibrated on the accepted predictions from the current exam, ensuring the confidence gate remains well-aligned with the updated model. 

\section{Experiments}
\subsection{Datasets}\label{sec:data}
We evaluate \chill{} on three ASAG datasets spanning different domains, grading scales, and difficulty levels. Table~\ref{tab:dataset_stats} summarizes their key characteristics.
\paragraph{DAMI.}
This dataset contains 4,031 anonymized student responses from a second-cycle Data Mining course, graded on multiple scales ($G$~$\in\{5,8,10\}$). We split the data into training (3,770 samples, 53 questions), calibration (260 samples, 12 questions), and test (261 samples, 53 questions). The test set covers unseen answers~(UA; 177 samples, 39 questions) and unseen questions~(UQ; 84 samples, 14 questions). For HiL experiments, $\mathcal{D}_\text{test}$ is divided into $\mathcal{D}_{21}$ (130 samples) for collecting corrections and $\mathcal{D}_{22}$ (131 samples) for post‑adaptation evaluation. The dataset’s heterogeneous rubrics and question diversity make it representative of real exam‑grading conditions.

\paragraph{SciEntsBank.}
SciEntsBank\footnote{https://huggingface.co/datasets/nkazi/SciEntsBank}~\cite{dzikovska-etal-2013-semeval} offers 10,804 elementary‑level science answers (grades 3–6) graded on a 0–4 scale ($G$~$\in\{0,1,2,3,4\}$). Following the standard setup, we evaluate in the fully UQ setting: training uses 4,969 responses from 135 questions, calibration draws 540 responses from training questions, and testing uses 733 responses from 30 unseen questions. This protocol imposes strict generalization demands, requiring models to handle entirely novel question types.

\paragraph{EngSAF.}
The EngSAF dataset\footnote{https://github.com/dishankaggarwal/EngSAF}~\cite{aggarwal2025understand} includes 5,798 responses across 119 questions from 25 engineering courses, graded on a 3‑point scale ($G$~$\in\{0,1,2\}$). We adopt the standard split: 3,650 training samples, 405 calibration samples, and 1,743 test samples, comprising both unseen answers (980) and unseen questions (763). EngSAF’s multi‑course coverage and coarser rubric assess whether methods generalize beyond single‑course, fine‑grained scoring.

\begin{table}[t]
\centering
\setlength{\tabcolsep}{4pt}
\begin{tabular}{lcccccc}
\toprule
\textbf{Dataset} & \textbf{Domain} & \textbf{Train} & \textbf{Cal} & \textbf{Test} & \textbf{MaxGrade} & \textbf{Eval Type} \\
\midrule
DAMI        & Data Mining  & 3,770 & 260 & 261  & 5/8/10 & UA + UQ \\
SciEntsBank & Science (K--6)  & 4,969 & 540 & 733  & 4        & UQ \\
EngSAF      & Engineering     & 3,650 & 405 & 1,743 & 2       & UA + UQ \\
\bottomrule
\end{tabular}
\smallskip \smallskip
\caption{Dataset statistics. MaxGrade indicates the grading scale. UA = Unseen Answers (questions in train), UQ = Unseen Questions (questions not in train).}
\label{tab:dataset_stats}
\end{table}
In all experiments, HiL corrections are simulated using ground‑truth grades for each split, enabling controlled evaluation of routing quality and adaptation effects without additional instructor effort. Deploying \chill{} with active instructors requires no changes to the framework.

\subsection{Setup}

\paragraph{Model and Hardware Configuration} We use Qwen-2.5-7B-Instruct~\cite{qwen2_2024} as our base model, selected for its strong instruction-following capabilities and consistent performance across diverse tasks. The model is fine-tuned using LoRA~\cite{hu2022lora} with ($r=16$, $\alpha=32$, dropout $0.1$) using AdamW optimizer with learning rate $2 \times 10^{-4}$, effective batch size 64 (batch size $2$, gradient accumulation $32$), and 6 epochs with linear warmup over the first $10\%$ of the steps. All experiments are conducted on 2$\times$ NVIDIA RTX A5500 GPUs (24\,GB each). Initial instruction-tuning on the DAMI (3,770 samples) completes in approximately 90 minutes. Each subsequent HiL cycle comprising calibration, selective prediction, adapter fine-tuning, and recalibration scales with rejected samples rather than the full dataset size, keeping per-cycle cost proportional to the teacher reviews, making \chill{} practical for real-world deployment.

\paragraph{Evaluation Metric} We evaluate grading performance using Quadratic Weighted Kappa~(QWK)~\cite{cohen1968weighted}, which measures ordinal agreement between predicted and reference grades while penalizing larger disagreements quadratically:
\begin{equation}
    \text{QWK} = 1 - \frac{\sum_{g,h \in \mathcal{G}} w_{gh}\, O_{gh}}
    {\sum_{g,h \in \mathcal{G}} w_{gh}\, E_{gh}},
    \label{eq:qwk}
\end{equation}
where $O_{gh}$ is the observed count of predictions $\hat{g}=g$ when the true grade is $h$, $E_{gh}$ is the expected count under random agreement, and $w_{gh} = (g-h)^2$ assigns larger penalties to larger errors. QWK ranges from $-1$ (complete disagreement) to $1$ (perfect agreement), with values above $0.8$ indicating strong agreement, comparable to inter-rater reliability among human graders. For HIL evaluation, we report both \emph{full-set QWK} (all test samples) and \emph{accepted-set QWK} (auto-accepted samples) to separately characterize overall system performance and the quality achieved under selective predictions. 

\paragraph{Hyperparameters}
The confidence threshold $\tau$ controls the coverage–quality trade off. We sweep $\tau \in \{0.4, 0.5, 0.6, 0.8, 0.9\}$ on the held‑out calibration split and select the value that maximizes accepted‑set QWK while keeping coverage in a practical range (typically 35–65\%). For temperature calibration, we grid‑search $T \in [0.1, 2.0]$ with step size $0.001$, choosing the value that minimizes ECE with $B = 10$ bins; this procedure takes about 5 minutes per dataset. The replay‑buffer size $k=3$ is based on ablations: $k=1$ underperforms when the rejected set is very small (3–5 samples per iteration), whereas larger $k$ increases computation with little gain. For each question in the rejected set, we retrieve its $k$ most similar training questions using Sentence‑BERT~\cite{reimers2019sbert} embeddings for question‑level similarity.

\subsection{Comparison to Baselines}

We establish baseline performance on \emph{DAMI} by comparing prompt engineering with in-context learning, RAG, and LoRA-based instruction-tuning. All model and prompt selection decisions are made using \emph{DAMI}, while \emph{SciEntsBank} and \emph{EngSAF} are reserved for evaluating the generalization of \chill{}. Table~\ref{tab:baselines_dami} summarizes the best configuration per approach.

\begin{table}[t]
\centering
\begin{tabular}{lcc}
\toprule
\textbf{Method} & \textbf{Model} & \textbf{QWK} \\
\midrule
Zero-shot & Llama-3.1-8B & 0.289 \\
Few-shot (k=3) & Llama-3.1-8B & 0.587 \\
Few-shot (k=5) & Llama-3.1-8B & 0.603 \\
\midrule
RAG only & Llama-3.1-8B & 0.443 \\
RAG + Few-shot (k=3) & Llama-3.1-8B & 0.491 \\
\midrule
\multirow{3}{*}{Fine-tuning (LoRA)} & Llama-3.1-8B & 0.677\textsuperscript{†} \\
 & Llama-3.2-3B & 0.547 \\
 & Qwen-2.5-7B & 0.669 \\
\midrule
\chill & Qwen-2.5-7B & \textbf{0.882} \\
\bottomrule
\end{tabular}
\smallskip \smallskip
\caption{Baseline results on DAMI. \textsuperscript{†}Llama-3.1-8B achieves the highest QWK but with severe systematic overgrading ($+1.87$ mean grade offset), disqualifying it for deployment. Fine-tuned models substantially outperform all prompt-based baselines.}
\label{tab:baselines_dami}
\end{table}
\paragraph{Prompting.} Zero-shot prompting achieves a QWK of only $0.289$, indicating that pretrained models cannot reliably grade student responses without task-specific adaptation. Adding five in-context examples ($k=5$) doubles the performance to $0.603$. However, gains plateau and reverse for $k > 5$, as context dilution degrades the model's ability to effectively follow instructions and grade consistently. We evaluated four prompt templates (basic, detailed, json\_strict, rubric\_strict; see Appendix) across six models (Llama-3.1-8B~\cite{llama3}, Llama-3.2-3B~\cite{llama3}, Qwen-2.5-7B~\cite{qwen2_2024}, Gemma-3-4B/12B~\cite{gemma2024}, Mistral-7B~\cite{mistral7b2023}), with Llama-3.1-8B using the basic template and $k=5$ achieving the strongest prompt-only result.

\paragraph{Retrieval Augmented Generation.}
We segment lecture notes into semantically coherent segments and retrieving the top-3 most similar chunks for each question using Sentence-BERT embeddings. Pure RAG achieves a QWK of $0.443$, underperforming few-shot ($k=5$) by 27\%. Combining retrieval with three in-context examples (RAG + few-shot, $k=3$) reaches a QWK of $0.491$, below both few-shot $k=3$ ($0.587$) and $k=5$ ($0.603$). Retrieval helps when the in-context budget is constrained, but does not close the gap to strong few-shot baselines, likely because retrieved lecture content improves factual recall without consistently aligning with rubric-level grading criteria.

\paragraph{Instruction-Tuning with LoRA.}
Instruction-tuning substantially outperforms prompt-based approaches. Among the three models evaluated (Qwen-2.5-7B, Llama-3.1-8B, and Llama-3.2-3B), Qwen-2.5-7B achieves $0.669$ QWK with a near-zero grade bias, making it the preferred model despite marginally lower QWK than Llama-3.1-8B ($0.677$). Llama-3.1-8B exhibits severe systematic overgrading, assigning grades $1.87$ above ground truth on average, a bias unacceptable in educational settings where grade integrity must be preserved. Llama-3.2-3B performs poorly ($0.547$ QWK) and is excluded from further experiments.

All fine-tuned models exhibit substantial validation-to-test gaps (mean $\Delta = 0.250$), revealing sensitivity to distribution shift even within the same course. This brittleness motivates \chill{}; rather than relying on static checkpoints that degrade over time, \chill{} enables continuous adaptation as the model encounters new question types and rubric patterns during deployment.

\subsection{Main Results}

\paragraph{Calibration (Problem~\ref{prob:minmax}).} Instruction-tuned models are severely overconfident prior to calibration, with baseline ECE reaching $0.271$ on \emph{DAMI}. Temperature scaling directly addresses this by rescaling confidence scores to match empirical accuracy, without altering the model's predictions or QWK. As shown in Table~\ref{tab:calibration}, \chill{} reduces ECE by 65\% on \emph{DAMI} and $53\%$ on \emph{EngSAF}, bringing both datasets into the well-calibrated zone ($\text{ECE} < 0.1$). \emph{SciEntsBank} exhibits only marginal improvement (6\% reduction), as the baseline model is already well-calibrated $\text{ECE} = 0.089$). The direction of the optimal temperature is itself informative. On \emph{DAMI}, it requires sharpening ($T^* = 0.337 < 1$), indicating that the base model produces conservative confidence estimates that understate its actual reliability, while on \emph{SciEntsBank} and \emph{EngSAF} datasets, the model requires smoothing ($T^* > 1$) to deflate overconfident predictions. This dataset-specific variation confirms that fixed temperature cannot generalize across grading domains, and that calibration on held-out data is a necessary component in a practical deployment pipeline.

\begin{table}[t]
\centering
\begin{tabular}{lcccc}
\toprule
\textbf{Dataset} & \textbf{Baseline} & \textbf{Calibrated} & \textbf{Optimal} & \textbf{ECE} \\
& \textbf{ECE} & \textbf{ECE} & \textbf{T*} & \textbf{Reduction} \\
\midrule
DAMI        & 0.270 & 0.094 & 0.337 & -65\% \\
SciEntsBank & 0.089 & 0.084 & 1.170 & -6\% \\
EngSAF      & 0.097 & 0.046 & 1.816 & -53\% \\
\bottomrule
\end{tabular}
\smallskip \smallskip
\caption{Calibration results. Temperature scaling reduces ECE by 53--65\% on DAMI and EngSAF while leaving QWK unchanged. SciEntsBank requires minimal adjustment as the base model is already near-calibrated.}
\label{tab:calibration}
\end{table}

\paragraph{HiL adaptation (Problem~\ref{prob:shift}).} Table~\ref{tab:hil_progression} reports HiL performance across correction cycles. On \textbf{\emph{DAMI}}, a single cycle raises QWK from $0.458$ to $0.882$ on $\mathcal{D}_{22}$ ($\Delta = +0.424$), surpassing the expert-level threshold $\tau=0.8$, while maintaining 35.1\% automated coverage. Thus, over a third of responses are graded automatically at expert-level quality, with the remainder routed to human review. On \textbf{\emph{SciEntsBank}}, the most challenging fully UQ condition, \chill{} improves progressively across three cycles, peaking at $0.764$ QWK at iteration 1 ($\Delta = +0.338$). The plateau at cycle 2 ($\Delta = +0.020$), reflects a temporary distribution shift in $\mathcal{D}_{23}$, followed by recovery to $0.715$ ($\Delta=+0.303$) QWK at cycle 3. This recovery illustrates the continual learning nature of \chill{}, preventing persistent degradation under rubric drift. On \textbf{\emph{EngSAF}}, the high baseline coverage (90--95\%) limits gains per cycle. However, the effect of threshold selection is most visible on $\mathcal{D}_{24}$, raising $\tau$ from $0.5$ to $0.8$ reduces coverage to 44.6\% while boosting QWK from $0.602$ to $0.840$ ($\Delta=+0.238$). This highlights $\tau$ as a critical parameter, allowing practitioners to balance automation and grading reliability without retraining. Finally, on \emph{DAMI}, \chill{}’s routing quality is evident from the gap between accepted and rejected predictions: the accepted 35\% achieve $0.882$ QWK, while rejected responses drop to $0.535$ QWK with a $3.1\times$ higher MAE. This $+0.347$ QWK difference shows that the confidence gate reliably separates trustworthy from uncertain cases.

\begin{table}[t]
\centering
\begin{tabular}{lccccccc}
\toprule
\textbf{\quad Dataset} & \textbf{Split} & \textbf{Iteration} & \textbf{Coverage} & \textbf{Baseline} & \textbf{Acc.} & \textbf{$\Delta$} & \textbf{Rej.} \\
 & & & \textbf{(\%)} & \textbf{QWK} & \textbf{QWK} & & \textbf{QWK} \\
\midrule
\multirow{2}{*}{\quad DAMI} & $\mathcal{D}_{21}$ & 0 & 97.7 & --- & 0.721 & --- & --- \\
 & $\mathcal{D}_{22}$ & 1 & 35.1 & 0.458 & \textbf{0.882} & +0.424 & 0.535 \\
\midrule
\multirow{4}{*}{\quad SciEntsBank} & $\mathcal{D}_{21}$ & 0 & 70.1 & --- & 0.436 & --- & 0.000 \\
 & $\mathcal{D}_{22}$ & 1 & 36.1 & 0.426 & \textbf{0.764} & +0.338 & 0.253 \\
 & $\mathcal{D}_{23}$ & 2 & 43.2 & 0.390 & 0.410 & +0.020 & 0.327 \\
 & $\mathcal{D}_{24}$ & 3 & 47.5 & 0.412 & 0.715 & +0.303 & 0.501 \\
\midrule
\multirow{5}{*}{\quad EngSAF} & $\mathcal{D}_{21}$ & 0 & 95.9 & --- & 0.589 & --- & 0.441 \\
 & $\mathcal{D}_{22}$ & 1 & 92.2 & 0.686 & 0.693 & +0.007 & 0.070 \\
 & $\mathcal{D}_{23}$ & 2 & 93.1 & 0.650 & 0.662 & $-$0.003 & 0.049 \\
 & $\mathcal{D}_{24}$ ($\tau$=0.5) & 3 & 90.3 & 0.584 & 0.672 & +0.076 & 0.085 \\
 & $\mathcal{D}_{24}$ ($\tau$=0.8) & 3 & 44.6 & 0.602 & \textbf{0.840} & +0.238 & 0.420 \\
\bottomrule
\end{tabular}
\smallskip \smallskip
\caption{Progressive HiL learning across correction cycles. Baseline QWK is computed on each split's accepted subset prior to corrections; Acc.\ QWK is computed after adapter fine-tuning; $\Delta$ = Acc.\ QWK $-$ Baseline QWK. Rej.\ QWK is the model's performance on human-routed samples.}
\label{tab:hil_progression}
\end{table}

\section{Conclusion}

We introduced \chill{}, a calibrated human-in-the-loop framework for short answer grading that addresses two core challenges in LLM‑based grading:  overconfidence and performance degradation under distribution shift. By post-hoc \textbf{temperature scaling}, confidence-based \textbf{selective prediction}, and \textbf{continual learning}, \chill{} achieves expert-level performance, consistently at $\approx0.80$ QWK on $35-65\%$ of student responses while deferring uncertain and low confidence cases to human review. Across three datasets in computer science, natural science, and engineering, our results highlight that calibration is essential: uncalibrated models exhibit unreliable confidence, but temperature scaling consistently achieves $\text{ECE} \le 0.1$. The optimal temperature is dataset‑dependent and must be updated as data evolves. Replay augmentation also proves critical, as removing it reduces the QWK to $0.025$ due to catastrophic forgetting. Unlike static grading systems, \chill{} adapts through rapid correction cycles, enabling practical deployment as rubrics and questions shift. By combining calibration, selective prediction, and continual learning, \chill{} offers reliable partial automation—handling high‑confidence cases automatically while keeping instructors in the loop for the rest. Future work includes evaluating \chill{} in live examination settings, studying multi‑instructor workflows, and extending the framework to multi‑modal responses (e.g., diagrams or code executions). Another promising direction is exploring adaptive gating policies that jointly optimize workload and accuracy, and integrating richer forms of feedback to further accelerate continual learning.

\begin{credits}
\subsubsection{\ackname} This research is funded by the European Health and Digital Executive Agency (HADEA) for the project AI and Health (Grant Agreement 101083880). Views and opinions expressed are, however, those of the author(s) only and do not necessarily reflect those of the European Union or the European Research Executive Agency. \euflag
\end{credits}

\bibliographystyle{splncs04}
\bibliography{references}

@article{marginson2016worldwide,
  author    = {Simon Marginson},
  title     = {The worldwide trend to high participation higher education: Dynamics of social stratification in inclusive systems},
  journal   = {Higher Education},
  year      = {2016},
  volume    = {72},
  number    = {4},
  pages     = {413--434},
  doi       = {10.1007/s10734-016-0016-x}
}

@inproceedings{aggarwal2025understand,
  title={“I Understand Why I Got This Grade”: ASAG with Feedback},
  author={Aggarwal, Dishank and Sil, Pritam and Raman, Bhaskaran and Bhattacharyya, Pushpak},
  booktitle={International Conference on Artificial Intelligence in Education},
  pages={304--318},
  year={2025},
  organization={Springer Nature}
}

@inproceedings{duong2024asag,
  author={Duong, Ta Nguyen Binh and Meng, Chai Yi},
  booktitle={2024 IEEE Global Engineering Education Conference}, 
  title={Automatic Grading of Short Answers Using Large Language Models in Software Engineering Courses}, 
  year={2024},
  volume={},
  number={},
  pages={1-10},
  doi={10.1109/EDUCON60312.2024.10578839}}

@misc{zhang2025it,
      title={Instruction Tuning for Large Language Models: A Survey}, 
      author={Shengyu Zhang and Linfeng Dong and Xiaoya Li and Sen Zhang and Xiaofei Sun and Shuhe Wang and Jiwei Li and Runyi Hu and Tianwei Zhang and Fei Wu and Guoyin Wang},
      year={2025},
      eprint={2308.10792},
      archivePrefix={arXiv},
      primaryClass={cs.CL},
      url={https://arxiv.org/abs/2308.10792}, 
}

@inproceedings{posocco2021calibrate,
    author="Posocco, Nicolas and Bonnefoy, Antoine",
    title="Estimating Expected Calibration Errors",
    booktitle="Artificial Neural Networks and Machine Learning",
    year="2021",
    publisher="Springer International Publishing",
    address="Cham",
    pages="139--150",
    year = "2021",
    isbn="978-3-030-86380-7"
}

@inproceedings{reimers2019sbert,
  title = "Sentence-BERT: Sentence Embeddings using Siamese BERT-Networks",
  author = "Reimers, Nils and Gurevych, Iryna",
  booktitle = "Conference on Empirical Methods in Natural Language Processing",
  month = "11",
  year = "2019",
  publisher = "ACL",
  url = "https://arxiv.org/abs/1908.10084",
}

@article{mistral7b2023,
  title        = {Mistral 7B},
  author       = {Jiang, Albert Q. and Sablayrolles, Alexandre and Roux, Antoine and others},
  year         = {2023},
  journal      = {arXiv preprint arXiv:2310.06825},
  url          = {https://arxiv.org/abs/2310.06825}
}

@article{gemma2024,
  title        = {Gemma: Open Models Based on Gemini Research and Technology},
  author       = {Google DeepMind},
  year         = {2024},
  journal      = {arXiv preprint arXiv:2403.08295}
}

@article{cohen1968weighted,
  title   = {Weighted Kappa: Nominal Scale Agreement with Provision for Scaled Disagreement or Partial Credit},
  author  = {Cohen, Jacob},
  journal = {Psychological Bulletin},
  volume  = {70},
  number  = {4},
  pages   = {213--220},
  year    = {1968},
  doi     = {10.1037/h0026256}
}

@article{qwen2_2024,
  title        = {Qwen2 Technical Report},
  author       = {Qwen Team, Alibaba Group},
  year         = {2024},
  journal      = {arXiv preprint arXiv:2407.10671},
  url          = {https://arxiv.org/abs/2407.10671}
}

@misc{llama3,
      title={The Llama 3 Herd of Models}, 
      author={Aaron Grattafiori and et al.},
      year={2024},
      eprint={2407.21783},
      archivePrefix={arXiv},
      primaryClass={cs.AI},
      url={https://arxiv.org/abs/2407.21783}, 
}

@inproceedings{feng2023aesqwk,
     author = {Afrizal Doewes and Nughthoh Arfawi Kurdhi and Akrati Saxena},
     booktitle = {16th International Conference on Educational Data Mining},
     doi = {10.5281/zenodo.8115784},
     isbn = {978-1-7336736-4-8},
     month = {July},
     pages = {103--113},
     publisher = {International Educational Data Mining Society},
     title = {Evaluating Quadratic Weighted Kappa as the Standard Performance Metric for Automated Essay Scoring},
     year = {2023}
}

@misc{nixon2020measuring,
title={Measuring Calibration in Deep Learning},
author={Jeremy Nixon and Mike Dusenberry and Ghassen Jerfel and Linchuan Zhang and Dustin Tran},
year={2020},
url={https://openreview.net/forum?id=r1la7krKPS}
}

@inproceedings{jimmy2021asag,
    author="Ljungman, Jimmy
    and Lislevand, Vanessa
    and Pavlopoulos, John
    and Farazouli, Alexandra
    and Lee, Zed
    and Papapetrou, Panagiotis
    and Fors, Uno",
    title="Automated Grading of Exam Responses: An Extensive Classification Benchmark",
    booktitle="Discovery Science",
    year="2021",
    publisher="Springer International Publishing",
    address="Cham",
    pages="3--18",
    isbn="978-3-030-88942-5"
}

@inproceedings{xie2024calibrating,
    title = "Calibrating Language Models with Adaptive Temperature Scaling",
    author = "Xie, Johnathan  and
      Chen, Annie S  and
      Lee, Yoonho  and
      Mitchell, Eric  and
      Finn, Chelsea",
    booktitle = "Proceedings of EMNLP 2024: System Demonstrations",
    year = "2024",
    publisher = "ACL",
    url = "https://aclanthology.org/2024.emnlp-main.1007/",
    doi = "10.18653/v1/2024.emnlp-main.1007",
    pages = "18128--18138"
}

@inproceedings{ovadia2019trust,
    author = {Ovadia, Yaniv and Fertig, Emily and Ren, Jie and Nado, Zachary and Sculley, D. and Nowozin, Sebastian and Dillon, Joshua V. and Lakshminarayanan, Balaji and Snoek, Jasper},
    title = {Can you trust your model's uncertainty? evaluating predictive uncertainty under dataset shift},
    year = {2019},
    publisher = {Curran Associates Inc.},
    booktitle = {Advances in NeurIPS},
    articleno = {1254},
    numpages = {12}
}

@article{delange2022continual,
  author    = {De Lange, Matthias and Aljundi, Rahaf and Masana, Marc and Parisot, Sarah and Jia, Xu and Leonardis, Ale{\v{s}} and Slabaugh, Gregory and Tuytelaars, Tinne},
  title     = {A Continual Learning Survey: Defying Forgetting in Classification Tasks},
  journal   = {IEEE Transactions on Pattern Analysis and Machine Intelligence},
  volume    = {44},
  number    = {7},
  pages     = {3366--3385},
  year      = {2022},
  publisher = {IEEE}
}

@article{burrows2015eras,
  author  = {Burrows, Steven and Gurevych, Iryna and Stein, Benno},
  title   = {The Eras and Trends of Automatic Short Answer Grading},
  journal = {International Journal of Artificial Intelligence in Education},
  volume  = {25},
  number  = {1},
  pages   = {60--117},
  year    = {2015},
  publisher = {Springer}
}

@inproceedings{geng2024survey,
  author    = {Geng, Jiahui and Cai, Fengyu and Wang, Yuxia and Koeppl, Heinz and Nakov, Preslav and Gurevych, Iryna},
  title     = {A Survey of Confidence Estimation and Calibration in Large Language Models},
  booktitle = {NAACL},
  pages     = {6577--6595},
  year      = {2024},
  publisher = {ACL}
}

@inproceedings{xiong2024llms,
  author    = {Xiong, Miao and Hu, Zhiyuan and Lu, Xinyang and Li, Yifei and Fu, Jie and He, Junxian and Hooi, Bryan},
  title     = {Can LLMs Express Their Uncertainty? An Empirical Evaluation of Confidence Elicitation in LLMs},
  booktitle = {12th ICLR},
  year      = {2024}
}

@inproceedings{wang2023orthogonal,
  author    = {Wang, Xiao and Chen, Tianze and Ge, Qiming and Xia, Han and Bao, Rong and Zheng, Rui and Zhang, Qi and Gui, Tao and Huang, Xuanjing},
  title     = {Orthogonal Subspace Learning for Language Model Continual Learning},
  booktitle = {Findings of the ACL: EMNLP 2023},
  pages     = {10658--10671},
  year      = {2023}
}

@inproceedings{hu2022lora,
    title={Lo{RA}: Low-Rank Adaptation of Large Language Models},
    author={Edward J Hu and yelong shen and Phillip Wallis and Zeyuan Allen-Zhu and Yuanzhi Li and Shean Wang and Lu Wang and Weizhu Chen},
    booktitle={International Conference on Learning Representations},
    year={2022},
    url={https://openreview.net/forum?id=nZeVKeeFYf9}
}

@inproceedings{guo2017calibration,
    author = {Guo, Chuan and Pleiss, Geoff and Sun, Yu and Weinberger, Kilian Q.},
    title = {On calibration of modern neural networks},
    year = {2017},
    publisher = {JMLR.org},
    booktitle = {Proceedings of the 34th International Conference on Machine Learning},
    pages = {1321–1330},
    numpages = {10}
}

@inproceedings{kotha2024understanding,
    title={Understanding Catastrophic Forgetting in Language Models via Implicit Inference},
    author={Suhas Kotha and Jacob Mitchell Springer and Aditi Raghunathan},
    booktitle={International Conference on Learning Representations},
    year={2024},
    url={https://openreview.net/forum?id=VrHiF2hsrm}
}

@article{luo2025forget,
  author={Luo, Yun and Yang, Zhen and Meng, Fandong and Li, Yafu and Zhou, Jie and Zhang, Yue},
  journal={IEEE Transactions on Audio, Speech and Language Processing}, 
  title={An Empirical Study of Catastrophic Forgetting in Large Language Models During Continual Fine-Tuning}, 
  year={2025},
  volume={33},
  number={},
  pages={3776-3786},
  doi={10.1109/TASLPRO.2025.3606231}}

@article{
   Impey_2025,
   title={Using Large Language Models for Automated Grading of Student Writing about Science},
   ISSN={1560-4306},
   url={http://dx.doi.org/10.1007/s40593-024-00453-7},
   DOI={10.1007/s40593-024-00453-7},
   journal={International Journal of Artificial Intelligence in Education},
   publisher={Springer Science and Business Media LLC},
   author={Impey, Chris and Wenger, Matthew and Garuda, Nikhil and Golchin, Shahriar and Stamer, Sarah},
   year={2025},
   month=jan
}

@misc{zeinalipour2025,
      title={Advancing Student Writing Through Automated Syntax Feedback}, 
      author={Kamyar Zeinalipour and Mehak Mehak and Fatemeh Parsamotamed and Marco Maggini and Marco Gori},
      year={2025},
      eprint={2501.07740},
      archivePrefix={arXiv},
      primaryClass={cs.CL},
      url={https://arxiv.org/abs/2501.07740}, 
}

@misc{chu2025enhancingllmbasedshortanswer,
      title={Enhancing LLM-Based Short Answer Grading with Retrieval-Augmented Generation}, 
      author={Yucheng Chu and Peng He and Hang Li and Haoyu Han and Kaiqi Yang and Yu Xue and Tingting Li and Joseph Krajcik and Jiliang Tang},
      year={2025},
      eprint={2504.05276},
      archivePrefix={arXiv},
      primaryClass={cs.CL},
      url={https://arxiv.org/abs/2504.05276}, 
}

@misc{gorbatovski2024,
      title={Reinforcement learning for question answering in programming domain using public community scoring as a human feedback}, 
      author={Alexey Gorbatovski and Sergey Kovalchuk},
      year={2024},
      eprint={2401.10882},
      archivePrefix={arXiv},
      primaryClass={cs.CL},
      url={https://arxiv.org/abs/2401.10882}, 
}

@inproceedings{2025.EDM.short-papers.166,
 author = {Juliette Woodrow and Chris Piech and Sanmi Koyejo},
 booktitle = {International Conference on Educational Data Mining},
 doi = {10.5281/zenodo.15870266},
 isbn = {978-1-7336736-6-2},
 pages = {442--449},
 publisher = {International Educational Data Mining Society},
 title = {Improving Generative AI Student Feedback: Direct Preference Optimization with Teachers in the Loop},
 year = {2025}
}

@InProceedings{avalon,
author="Armfield, Derek
and Chen, Eason
and Omonkulov, Asilbek
and Tang, Xinyi
and Lin, Jionghao
and Thiessen, Erik
and Koedinger, Kenneth",
title="Avalon: A Human-in-the-Loop LLM Grading System with Instructor Calibration and Student Self-assessment",
booktitle="International Conference on Artificial Intelligence in Education.",
year="2025",
publisher="Springer Nature",
pages="111--118",
isbn="978-3-031-99267-4"
}

@Inbook{Burstein1999,
author="Burstein, Jill
and Wolff, Susanne
and Lu, Chi",
title="Using Lexical Semantic Techniques to Classify Free-Responses",
bookTitle="Breadth and Depth of Semantic Lexicons",
year="1999",
publisher="Springer Netherlands",
address="Dordrecht",
pages="227--244",
isbn="978-94-017-0952-1",
doi="10.1007/978-94-017-0952-1_11",
url="https://doi.org/10.1007/978-94-017-0952-1_11"
}

@article{Leacock2003CraterAS,
  title={C-rater: Automated Scoring of Short-Answer Questions},
  author={Claudia Leacock and Martin Chodorow},
  journal={Computers and the Humanities},
  year={2003},
  volume={37},
  pages={389-405},
  url={https://api.semanticscholar.org/CorpusID:27443635}
}

@inproceedings{dzikovska-etal-2013-semeval,
    title = "{S}em{E}val-2013 Task 7: The Joint Student Response Analysis and 8th Recognizing Textual Entailment Challenge",
    author = "Dzikovska, Myroslava  and
      Nielsen, Rodney  and
      Brew, Chris  and
      Leacock, Claudia  and
      Giampiccolo, Danilo  and
      Bentivogli, Luisa  and
      Clark, Peter  and
      Dagan, Ido  and
      Dang, Hoa Trang",
    booktitle = "{S}em{E}val 2013 Workshop Proceedings",
    month = jun,
    year = "2013",
    publisher = "ACL",
    url = "https://aclanthology.org/S13-2045/",
    pages = "263--274"
}

@inproceedings{mohler-mihalcea-2009-text,
    title = "Text-to-Text Semantic Similarity for Automatic Short Answer Grading",
    author = "Mohler, Michael  and
      Mihalcea, Rada",
    booktitle = "Proceedings of {EACL}",
    year = "2009",
    publisher = "ACL",
    url = "https://aclanthology.org/E09-1065/",
    pages = "567--575"
}
\appendix

\section{Extended Baseline Results}
\label{app:baselines}

Table~\ref{tab:extended_baselines} reports prompt engineering 
results across five models and four templates on \emph{DAMI}. 
Llama-3.1-8B consistently outperforms smaller models, peaking 
at the basic template with $k=5$ in-context examples. Smaller 
models (Llama-3.2-3B, Gemma-3-4B) benefit from more structured prompts and peak at $k=1$--$3$, reflecting limited capacity  for long-context utilization. Mistral-7B shows poor few-shot scaling and is excluded from further experiments.

\begin{table}[h]
\centering
\setlength{\tabcolsep}{4pt}
\begin{tabular}{llcccc}
\toprule
\textbf{Model} & \textbf{Prompt} & \textbf{Zero} & \textbf{FS-1} 
& \textbf{FS-3} & \textbf{FS-5} \\
\midrule
\multirow{4}{*}{Llama-3.1-8B} 
  & basic        & 0.289 & 0.549 & 0.587 & \textbf{0.603} \\
  & detailed     & 0.477 & 0.482 & 0.537 & 0.524 \\
  & json\_strict & 0.316 & 0.490 & 0.560 & 0.541 \\
  & rubric       & 0.425 & 0.532 & 0.507 & 0.506 \\
\midrule
\multirow{4}{*}{Llama-3.2-3B}
  & basic        & 0.256 & 0.383 & 0.324 & 0.337 \\
  & detailed     & 0.365 & \textbf{0.416} & 0.388 & 0.355 \\
  & json\_strict & 0.263 & 0.400 & 0.354 & 0.350 \\
  & rubric       & 0.290 & 0.377 & 0.344 & 0.316 \\
\midrule
\multirow{2}{*}{Gemma-3-4B}
  & basic        & 0.265 & 0.369 & 0.329 & 0.355 \\
  & detailed     & 0.368 & 0.383 & \textbf{0.395} & 0.389 \\
\midrule
Gemma-3-12B & basic & 0.330 & 0.462 & 0.433 & \textbf{0.507} \\
\midrule
Mistral-7B  & basic & 0.291 & \textbf{0.337} & 0.306 & 0.314 \\
\bottomrule
\end{tabular}
\smallskip
\caption{Full prompt engineering results on \emph{DAMI}. Bold indicates the best configuration per model. }
\label{tab:extended_baselines}
\end{table}

\begin{figure}[h]
    \centering
    \includegraphics[width=0.85\linewidth]{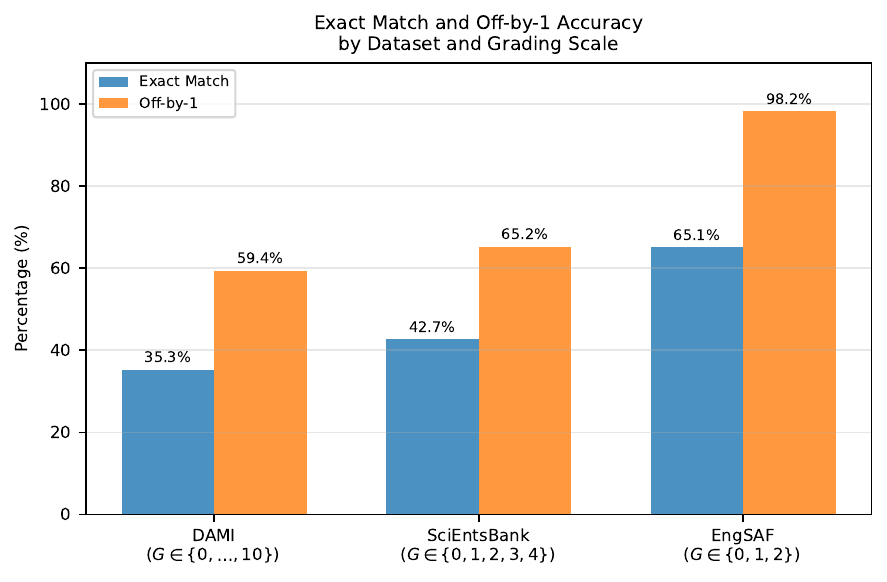}
    \caption{Exact Match and Off-by-1 accuracy for Qwen-2.5-7B across \emph{DAMI}, \emph{SciEntsBank}, and \emph{EngSAF}. The improvement across datasets reflects scale granularity rather than model quality, Off-by-1 on \emph{EngSAF} ($G \in \{0,1,2\}$) is a much coarser tolerance than on \emph{DAMI} ($G \in \{0,\ldots,10\}$).}
    \label{fig:em_offby1}
\end{figure}

\paragraph{Error analysis.}
Figure~\ref{fig:em_offby1} illustrates Exact Match (EM) and Off-by-1 accuracy for Qwen-2.5-7B across all three datasets. Off-by-1 measures the proportion of predictions within one grade point of the ground truth, capturing near-miss errors that may still be acceptable in practice. While EM and Off-by-1 improve progressively from \emph{DAMI} to \emph{EngSAF}, this progression is driven by scale granularity rather than model improvement. On \emph{EngSAF} ($G \in \{0,1,2\}$) almost any grading error qualifies as off-by-1, whereas on \emph{DAMI}  ($G \in \{0,\ldots,10\}$) being within one point is a strict tolerance. All EM and Off-by-1 numbers must therefore be interpreted within their respective grading scales. Among the fine-tuned models evaluated on \emph{DAMI}, Qwen-2.5-7B is the only model with near-zero systematic offset ($+0.03$), making it the only viable candidate for deployment, both the Llama variants exhibited severe overgrading bias, with the Llama-3.2-3B assigning grades $1.87$ points above ground truth on average.

\begin{figure}[ht]
    \centering
    \includegraphics[width=1\linewidth]{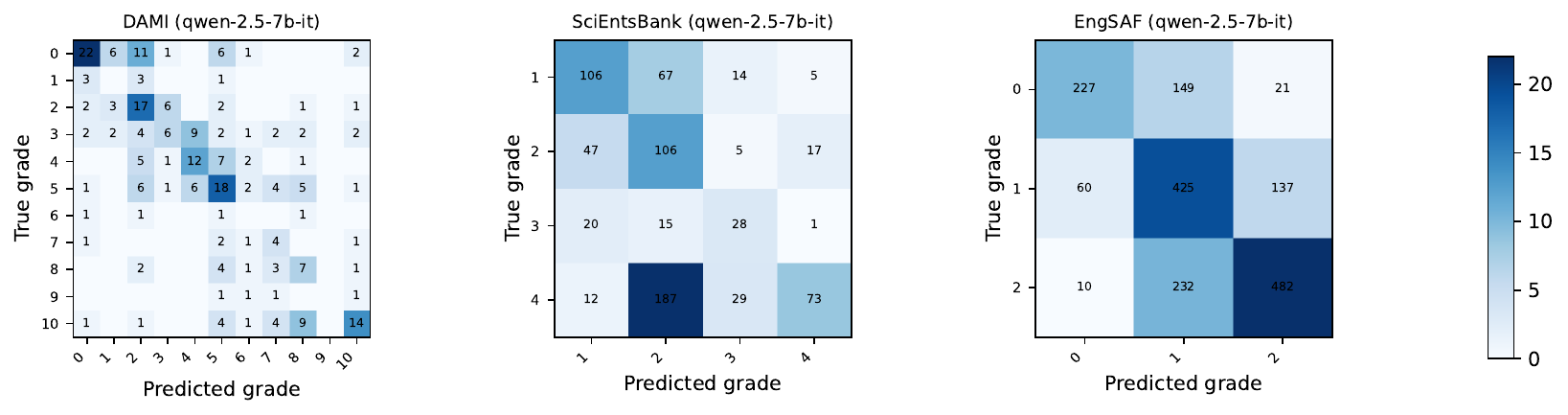}
    \caption{Confusion matrices for baseline models on \emph{DAMI}, \emph{SciEntsBank}, and \emph{EngSAF}. \emph{DAMI} errors concentrate within $\pm 1$ of the diagonal; \emph{SciEntsBank} shows systematic under-prediction of grade 4; \emph{EngSAF} confirms the relative 
    simplicity of 3-way grading.}
    \label{fig:confusion-matrix}
\end{figure}

\paragraph{Prediction quality.}
Figure~\ref{fig:confusion-matrix} shows the confusion matrices for baseline models across all three datasets. \emph{DAMI's} sparse distribution reflects its variable grading scales (0--10), with most errors falling within $\pm1$ grade of the diagonal, consistent with rubric-level near misses rather than severe mispredictions. Meanwhile, on the \emph{SciEntsBank}, the model exhibits systematic under-prediction of grade 4, suggesting that the model struggles to distinguish high quality responses from adequate ones. \emph{EngSAF's} concentrated diagonal, especially the 425 correct predictions for grade 1, confirms the relative simplicity of the 3-way classification task, though non-trivial confusion between grades 0 and 1 persists. Across all three datasets, the error patterns are consistent with the distribution shift problem: the model's weaknesses are not random, making them well-suited for target corrections via HiL supervision.

\paragraph{Performance by question type.}
Table~\ref{tab:ua_uq_breakdown} breaks down performance on \emph{DAMI} by question type, Unseen Answers (UA), where the question appeared in training but the specific response is new, and Unseen Questions (UQ), where neither the question nor its responses were seen during training. The baseline model performs better on UA than UQ ($0.705$ vs $0.365$ QWK), confirming that generalization to entirely new question types is a hard problem. After one HiL 
correction cycle, both categories improve, with the UQ gaining a QWK of $+0.109$ ($0.365 \to 0.474$), a proportionally larger improvement than UA ($+0.014$), suggesting that human corrections are particularly effective at addressing the model's weaknesses on novel question types.

\begin{table}[h]
\centering
\begin{tabular}{llccccc}
\toprule
\textbf{Model} & \textbf{Type} & \textbf{$n$} & \textbf{QWK} & \textbf{EM (\%)} & \textbf{Off-by-1 (\%)} & \textbf{MAE} \\
\midrule
\multirow{2}{*}{Baseline}
 & UA      &  89 & 0.705 & \textbf{41.6} & 61.8 & \textbf{1.427} \\
 & UQ      &  42 & 0.365 & 19.0 & 47.6 & 2.310 \\
\midrule
\multirow{2}{*}{Post-HiL}
 & UA      &  89 & \textbf{0.719} & 30.3 & \textbf{66.3} & 1.472 \\
 & UQ      &  42 & \textbf{0.474} & \textbf{23.8} & 47.6 & \textbf{2.024} \\
\bottomrule
\end{tabular}
\smallskip
\caption{Performance breakdown by question type on \emph{DAMI}. UA = Unseen Answers (questions seen during training); UQ = Unseen Questions (questions not seen during training). Baseline is the fine-tuned model on the full test set; Post-HiL is evaluated on $\mathcal{D}_{22}$ after one correction cycle.}
\label{tab:ua_uq_breakdown}
\end{table}

\section{Coverage-Quality Analysis}
\label{app:coverage}

\begin{figure}[h]
    \centering
    \includegraphics[width=\linewidth]{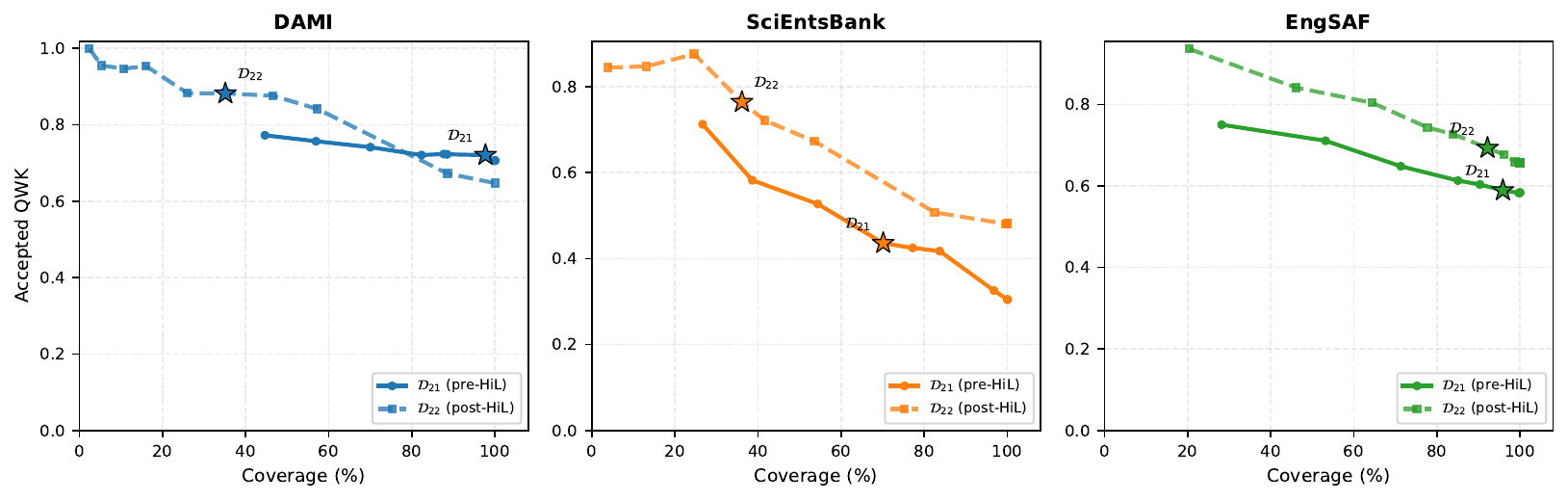}
    \caption{Coverage--quality curves for \emph{DAMI}, \emph{SciEntsBank}, 
    and \emph{EngSAF}. Each point corresponds to a specific $\tau$ 
    value; selected operating points are marked. Restricting 
    coverage to high-confidence predictions consistently 
    improves accepted-set QWK across all three datasets.}
    \label{fig:coverage_qwk}
\end{figure}

\paragraph{Coverage-quality tradeoff.}
Figure~\ref{fig:coverage_qwk} shows accepted-set QWK as a 
function of coverage across all three datasets, obtained by 
sweeping $\tau$ before and after HiL adaptation. Across all 
datasets, the post-HiL curve ($\mathcal{D}_{22}$, dashed) 
lies consistently above the pre-HiL curve ($\mathcal{D}_{21}$, solid), confirming that HiL adaptation improves grading quality at every operating point, not just at the selected threshold. \emph{SciEntsBank} exhibits the steepest pre-HiL quality gain with coverage, reflecting high variance in prediction confidence across its diverse question types; the post-HiL curve shows a pronounced upward shift, particularly at moderate coverage ($40$--$70\%$). On \emph{DAMI}, the post-HiL curve reaches $0.882$ at the selected operating point ($\mathcal{D}_{22}$, $35\%$ coverage), a substantial improvement over the pre-HiL curve at the same coverage level. \emph{EngSAF} shows a more modest but consistent upward shift, reflecting the smaller marginal gains available on its simpler 3-way grading scale.

\paragraph{Rejected samples quality gap.}
The confidence gate's discriminative quality is validated by
the gap between accepted and rejected predictions on \emph{DAMI}. The accepted subset ($35.1\%$ coverage) achieves a QWK of $0.882$, off-by-1 of $89.1\%$, and MAE $0.696$; the rejected subset scores $0.535$ QWK, $44.7\%$ off-by-1, and MAE $2.165$, $3.1\times$ worse. Of the $85$ rejected samples, $81.2\%$ carry wrong predictions (correct gate decisions), while the remaining $18.8\%$ are correct predictions that were
conservatively over-rejected. Breaking the rejected set down
by model uncertainty type: $40.0\%$ fall in the UQ category and $60.0\%$ in UA. This breakdown confirms that the confidence gate routes for the right reasons, and that the $+0.347$ QWK gap reflects genuine discriminative signal.

\section{Ablation Studies}
\label{app:ablations}

Table~\ref{tab:hil_failures} documents four design configurations on \emph{DAMI} that informed the final \chill{} setup. Scenario~1 shows that a 3B parameter model lacks the capacity to retain prior grading knowledge under fine-tuning, with QWK collapsing to $-0.071$ on $\mathcal{D}_{22}$ (worse than random). Scenario~2 shows that fine-tuning without a replay buffer causes catastrophic forgetting even in an 8B model, reducing QWK from $0.863$ to $0.025$ despite strong initial performance. Scenario~3 confirms that threshold selection is as critical as model and buffer design: $\tau=0.5$ maintains high quality ($0.916$) but at the cost of coverage collapsing from $26.2\%$ to $17.6\%$. Scenario~4 is the optimal configuration, reported in the main paper, where lowering $\tau$ to $0.4$ recovers practical coverage while achieving $0.882$ QWK on $\mathcal{D}_{22}$.

\begin{table}[h]
\centering
\resizebox{\textwidth}{!}{
\begin{tabular}{clccccc}
\toprule
\textbf{\#} & \textbf{Model} & \textbf{$\tau$} & \textbf{Replay} & \textbf{QWK ($\mathcal{D}_{21}$)} 
& \textbf{QWK ($\mathcal{D}_{22}$)}
& \textbf{Coverage ($\mathcal{D}_{22}$)} \\
\midrule
1 & Llama-3.2-3B & 0.5 & \checkmark & 0.854 & $-0.071$ & 10.7\% \\
2 & Llama-3.1-8B & 0.5 & $\times$   & 0.863 & 0.025    & 39.7\% \\
3 & Qwen-2.5-7B  & 0.5 & \checkmark & 0.925 & 0.916    & 17.6\% \\
4 & Qwen-2.5-7B  & 0.4 & \checkmark & 0.721 & 0.882    & 35.1\% \\
\bottomrule
\end{tabular}}
\smallskip
\caption{Design configurations evaluated on \emph{DAMI} leading to the final \chill{} setup. Each row isolates one design decision: model capacity (1), replay buffer (2), and threshold selection (3 vs 4). Scenario~4 is the configuration reported in the main paper.}
\label{tab:hil_failures}
\end{table}

\section{HiL Progression}
\label{app:hil}

\begin{figure}[h]
    \centering
    \includegraphics[width=0.85\linewidth]{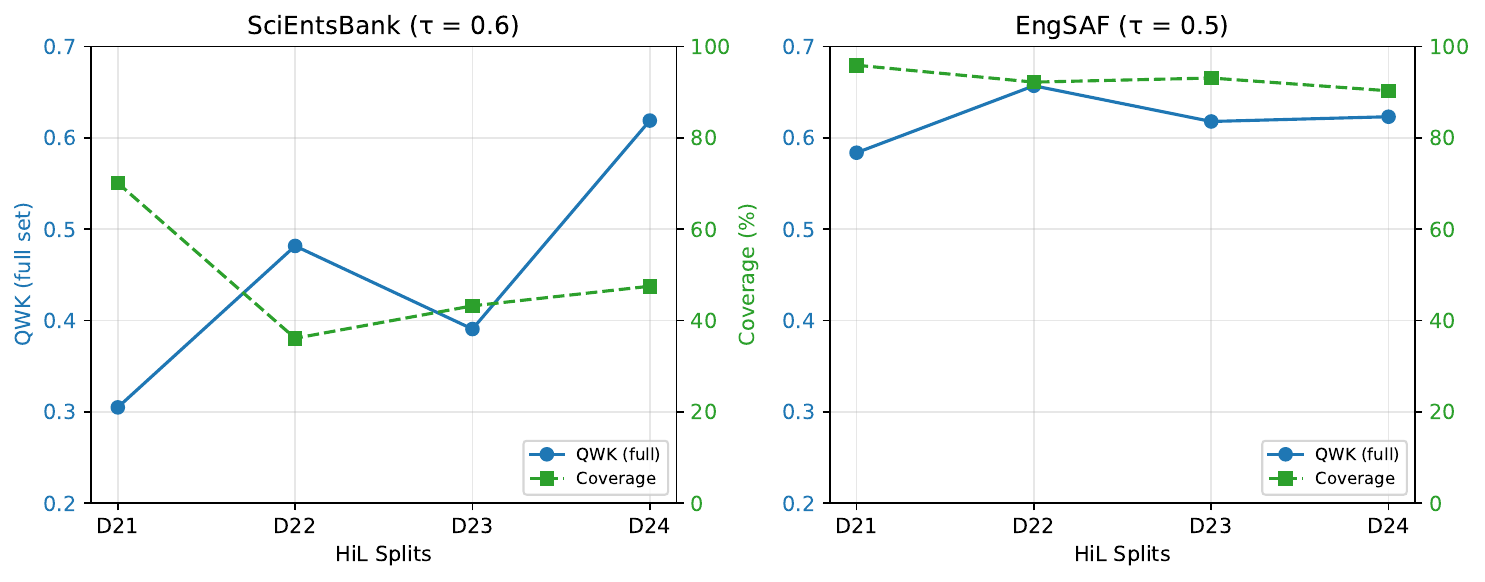}
    \caption{HiL progression across four correction cycles for 
    \emph{SciEntsBank} (left) and \emph{EngSAF} (right). Full-set QWK (blue, 
    left axis) and automated coverage (green dashed, right axis) 
    are shown at each split. \emph{SciEntsBank} shows a non-monotonic 
    but ultimately recovering QWK trajectory; \emph{EngSAF} maintains 
    high coverage throughout with steady QWK improvement.}
    \label{fig:hil_progression}
\end{figure}

Figure~\ref{fig:hil_progression} shows the full HiL progression across four correction cycles for \emph{SciEntsBank} and \emph{EngSAF}, reporting full-set QWK and automated coverage at each split. On \emph{SciEntsBank}, full-set QWK improves from $0.305$ to $0.619$ across four cycles while coverage stabilizes around $40$--$50\%$, reflecting the model's growing confidence as corrections accumulate. On \emph{EngSAF}, full-set QWK improves steadily from $0.584$ to $0.623$ at consistently high coverage ($\approx90\%$), confirming that \chill{} adapts reliably even under a coarser 3-way grading scale.

\section{Grade Distribution}
\label{app:label_dist}

\begin{figure}[ht]
    \centering
    \includegraphics[width=0.85\linewidth]{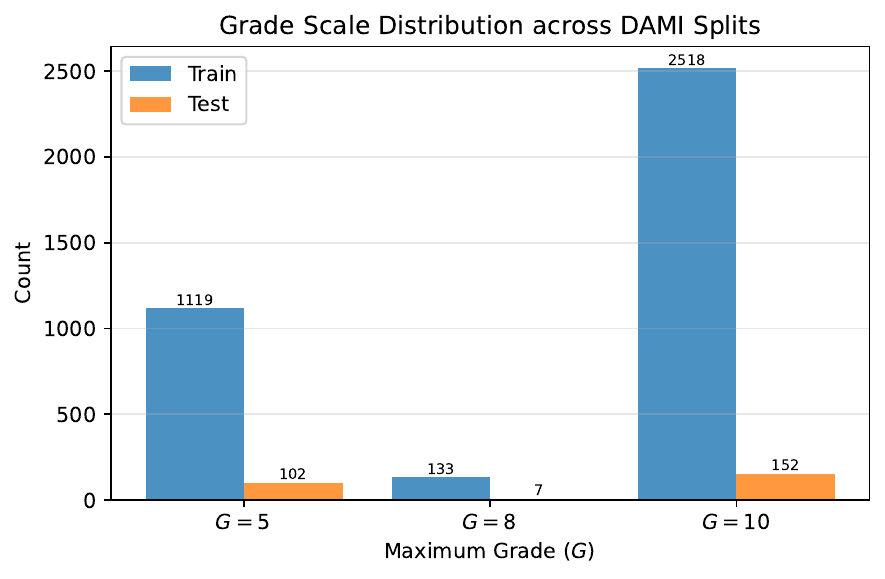}
    \caption{Grade scale distribution across train and test 
    splits of \emph{DAMI}. The $G=8$ scale is substantially 
    underrepresented ($133$ train, $7$ test samples), 
    motivating the scale-aware replay buffer in \chill{} 
    which explicitly balances representation across $G$ 
    categories during HiL adaptation.}
    \label{fig:label_dist}
\end{figure}

Figure~\ref{fig:label_dist} shows the grade scale distribution across train and test splits of \emph{DAMI}. The $G=10$ scale dominates, comprising $2{,}518$ training and $152$ test samples, while $G=8$ is substantially underrepresented with only $133$ training and $7$ test samples. This imbalance across scales directly motivates the scale-aware replay buffer in \chill{}, which explicitly balances representation across $G$ categories during adaptation to prevent fine-tuning from being dominated by the most frequent scale.

\section{Grade Granularity}
\label{app:normalized}

To assess whether collapsing \emph{DAMI's} heterogeneous grading scale to a much coarser 3-class scheme ($G \in \{0,1,2\}$) improves HiL performance, we train and evaluate \chill{} on a normalized \emph{DAMI} variant (Table~\ref{tab:grade_granularity}). The normalized model achieves a validation QWK of $0.768$, but a test QWK of just $0.430$, a significant val-to-test gap of $-0.338$, more 
than twice that of the original formulation ($-0.157$). 
Collapsing to 3 classes introduces borderline samples 
(e.g., grades 3/5 and 4/5 both mapping to grade~1) that 
are ambiguous at test time but appear consistent within 
training. Under \chill{} adaptation, post-fine-tuning temperature saturates at $T=1.969$, collapsing confidence scores and reducing $\mathcal{D}_{22}$ coverage to $4.6\%$ at 
$\tau=0.8$. Collapsing the rubric reduces the information available for calibrated routing, narrows the confidence distribution, and ultimately undermines the confidence gate. The original multi-scale formulation preserves the grade resolution that \chill{} depends on for reliable selective prediction.

\begin{table}[h]
\centering
\setlength{\tabcolsep}{4pt}
\begin{tabular}{lcc}
\toprule
& \textbf{Original scale} & \textbf{Normalized (3-class)} \\
\midrule
Val QWK                    & 0.826          & 0.768  \\
Test QWK                   & 0.705          & 0.430  \\
Val $\to$ Test gap         & $-$0.121       & $-$0.338 \\
$\mathcal{D}_{22}$ coverage     & 35.1\%    & 4.6\%  \\
$\mathcal{D}_{22}$ accepted QWK & \textbf{0.882} & 0.519 \\
\bottomrule
\end{tabular}
\smallskip
\caption{Grade granularity ablation on \emph{DAMI}. The normalized variant collapses to $G \in \{0,1,2\}$.}
\label{tab:grade_granularity}
\end{table}

\section{Prompt Templates}
\label{app:prompt}

Figures~\ref{fig:prompt_basic}--\ref{fig:prompt_rubric} show 
the four prompt templates evaluated in the baseline experiments 
(Table~\ref{tab:extended_baselines}). All templates share the 
same JSON output format and differ only in the amount of grading 
guidance provided. The basic template is used as the default 
for instruction-tuning and inference across all datasets.

\begin{figure}[h]
\centering
\begin{tcolorbox}[colback=gray!5, colframe=gray!40, width=\linewidth]
\small
\textbf{System:} You are an Automated Short Answer Grader (ASAG). 
Return ONLY a strict JSON with keys: \texttt{"grade"} (int), 
\texttt{"max\_grade"} (int).\\[4pt]
\textbf{User:}\\
Question: \textit{\{q\}}\\
Answer: \textit{\{a\}}\\
Target Scale: 0 to \textit{\{max\_g\}}
\end{tcolorbox}
\caption{Basic prompt template. Minimal context; used as the 
default for instruction-tuning and all cross-dataset experiments.}
\label{fig:prompt_basic}
\end{figure}

\begin{figure}[h]
\centering
\begin{tcolorbox}[colback=gray!5, colframe=gray!40, width=\linewidth]
\small
\textbf{System:} You are an expert educational assessment 
specialist. Grade student answers based on (1)~correctness, 
(2)~completeness, and (3)~clarity. Full marks for accurate and 
complete answers; proportional partial credit otherwise. Respond 
with ONLY a valid JSON object: 
\texttt{\{"grade": <int>, "max\_grade": <int>\}}\\[4pt]
\textbf{User:}\\
Question: \textit{\{q\}}\\
Student Answer: \textit{\{a\}}\\
Target Scale: 0 to \textit{\{max\_g\}} \\
(0 = completely incorrect; \textit{\{max\_g\}} = fully correct 
and complete)
\end{tcolorbox}
\caption{Detailed prompt template. Adds explicit grading criteria 
(correctness, completeness, clarity) and per-endpoint scale 
descriptors.}
\label{fig:prompt_detailed}
\end{figure}

\begin{figure}[h]
\centering
\begin{tcolorbox}[colback=gray!5, colframe=gray!40, width=\linewidth]
\small
\textbf{System:} You are a grading system that outputs ONLY 
valid JSON. Evaluate student answers based on correctness, 
completeness, and proportional partial credit. Required output 
(no exceptions): 
\texttt{\{"grade": <int between 0 and max\_grade>, 
"max\_grade": <int>\}}. 
No explanations. No commentary. Only JSON.\\[4pt]
\textbf{User:}\\
Question: \textit{\{q\}}\\
Student Answer: \textit{\{a\}}\\
Target Scale: 0 to \textit{\{max\_g\}}\\[2pt]
Output JSON:
\end{tcolorbox}
\caption{JSON-strict prompt template. Enforces strict output 
formatting with explicit constraints against free-text generation.}
\label{fig:prompt_json}
\end{figure}

\begin{figure}[h]
\centering
\begin{tcolorbox}[colback=gray!5, colframe=gray!40, width=\linewidth]
\small
\textbf{System:} You are an expert grader. Evaluate responses 
using this rubric: 100\% credit for fully correct and complete 
answers; 75--99\% for mostly correct with minor issues; 
50--74\% for partial understanding with notable gaps; 25--49\% 
for minimal correctness with major issues; 0--24\% for incorrect 
or irrelevant responses. Apply proportionally to the given scale. 
Return ONLY: 
\texttt{\{"grade": <int>, "max\_grade": <int>\}}\\[4pt]
\textbf{User:}\\
Question: \textit{\{q\}}\\
Student Answer: \textit{\{a\}}\\
Target Scale: 0 to \textit{\{max\_g\}}\\
(Apply the rubric framework proportionally to this scale)
\end{tcolorbox}
\caption{Rubric-based prompt template. Provides explicit 
percentage-based scoring descriptors applied proportionally 
to the grading scale.}
\label{fig:prompt_rubric}
\end{figure}

\section{Dataset Examples}
\label{app:examples}

Table~\ref{tab:dami_examples} presents representative examples from the \emph{DAMI} dataset, showing the range of question complexity, response quality, and grading subjectivity encountered in graduate-level short-answer assessment. The examples span multiple topics and grading scales, highlighting why reliable automated grading requires both calibrated confidence and human oversight for uncertain cases.

\begin{table}[h]
\centering
\footnotesize
\setlength{\tabcolsep}{4pt}
\begin{tabular}{p{5.0cm} p{5.0cm} c}
\toprule
\textbf{Question (truncated)} & \textbf{Student Response (truncated)} & \textbf{Grade} \\
\midrule
To what extent do the following methods allow constituent models to be generated in parallel: Bagging, Random Forests, Boosting, Stacking? &
Generated in parallel: Bagging and Random Forests. Some steps can be generated in parallel: Stacking. Not generated in parallel: Boosting. &
5/10 \\
\midrule
The standard K-Means algorithm loads all data into memory. Data instead arrives in a streaming manner. What modification would you suggest? &
Not necessary to modify the algorithm, just do each part individually with a constant stream of information. Use threads to treat, analyse, and send the data\ldots &
1/10 \\
\midrule
Consider a CNN with three consecutive 2$\times$2 conv.\ layers (stride=1, no pooling). How many original pixels activate a single neuron in the 2nd non-image layer? What if stride=2? &
Layer 1: $2{\times}2{=}4$. Striding right: $+2$, down: $+2$, diagonal: $+1$. Total $= 9$. With stride$=$2: $4{+}4{+}4{+}4{=}16$. &
5/5 \\
\midrule
Graph $G$ has 20 nodes and 190 edges (complete). Compute the indegree. What is the PageRank of any node? If half the edges are removed at random, what is the new PageRank? &
Indegree $= n{-}1 = 19$. Initial PageRank $= 1$ (complete graph). After removing half the edges, probability halves: PageRank $= 0.5$. &
2.5/5 \\
\bottomrule
\end{tabular}
\smallskip
\caption{Representative examples from the DAMI dataset illustrating the difficulty and subjectivity of graduate-level short-answer grading. Questions span multiple topics from a Master's-level data mining course; grades use either a 5-point or 10-point scale depending on question complexity.}
\label{tab:dami_examples}
\end{table}

\end{document}